\RequirePackage{fix-cm}
\documentclass{svjour3} 
\smartqed 
\usepackage{graphicx}
\usepackage{amsmath}
\usepackage{amsfonts}
\begin{document}

\title{Signalling Paediatric Side Effects using an Ensemble of Simple Study Designs}



\author{Jenna M. Reps \and
Jonathan M. Garibaldi \and Uwe Aickelin \and Daniele Soria \and Jack E. Gibson \and Richard B. Hubbard 
}


\institute{Jenna M. Reps \and Jonathan M. Garibaldi \and Uwe Aickelin \and Daniele Soria \at
IMA, The University of Nottingham,Nottingham, NG8 1BB \\
Tel.: +44 (0) 115 95 14299\\
\email{psxjr1@nottingham.ac.uk} 
\and
Jack E. Gibson \and Richard B. Hubbard \at
Community Health Sciences, Clinical Sciences Building, Nottingham, NG5 1PB
}

\date{Received: date / Accepted: date}

\maketitle

\begin{abstract}
Background: Children are frequently prescribed medication `off-label', meaning there has not been sufficient testing of the medication to determine its safety or effectiveness. The main reason this safety knowledge is lacking is due to ethical restrictions that prevent children from being included in the majority of clinical trials.\\

\noindent Objective: The objective of this paper is to investigate whether an ensemble of simple study designs can be implemented to signal acutely occurring side effects effectively within the paediatric population by using historical longitudinal data. The majority of pharmacovigilance techniques are unsupervised, but this research presents a supervised framework. \\

\noindent Methods: Multiple measures of association are calculated for each drug and medical event pair and these are used as features that are fed into a classifier to determine the likelihood of the drug and medical event pair corresponding to an adverse drug reaction. The classifier is trained using known adverse drug reactions or known non-adverse drug reaction relationships.\\

\noindent Results: The novel ensemble framework obtained a false positive rate of $0.149$, a sensitivity of  $0.547$ and a specificity of $0.851$ when implemented on a reference set of drug and medical event pairs. The novel framework consistently outperformed each individual simple study design. \\ 

\noindent Conclusion: This research shows that it is possible to exploit the mechanism of causality and presents a framework for signalling adverse drug reactions effectively.

\noindent \textbf {Key Points}\\
\begin{itemize}
\item The ensemble of simple measures outperformed each single simple measure when considering both the overall ability to rank adverse drug reactions and the signalling performance at a natural threshold.
\item The ensemble method is adaptable as it can incorporate any new measures of association that are propose over time.
\item The results of the paper highlight the potential benefit of applying supervised learning for adverse drug reaction signalling.
\end{itemize}
\end{abstract}

\section{Introduction} \label{sec:1}
There is an abundance of evidence to support the impression that side effects in children currently present a significant public health problem \cite{Aagaard2010,Conroy2000}. When there is a causal relationship between a drug and medical event it is termed an adverse drug reaction (ADR). Children of all ages can suffer from diseases that require them to take medication but the suitability of drugs in the paediatric population (0-17 years old) is generally unexplored. The majority of paediatric prescriptions are `off-label' meaning the licensed medication is used in situations that have not had sufficient investigation to determine the drug efficiency or safety. Examples of off-label prescriptions are prescribing a different dosage or frequency than recommended, prescribing a drug for a different indication than the drug was tested for or prescribing the drug to age groups such as children where the drug has not been extensively evaluated. The paediatric population are rarely involved in clinical trials, so there is little evidence available to determine if a medication is efficient and safe \cite{Rose2010} and many drugs do not have licenses for children. A study that observed a paediatric hospital in Derby, UK found that 23\% of the prescribed drugs were off-label and this was lower than the off-label rate observed in four other european hospitals \cite{Conroy2000}. 

The problem with `off-label' prescribing within the paediatric population is that there are clear physiological differences between adults and children, so the efficiency and safety knowledge discovered during clinical trials in the adult population cannot be accurately extrapolated for the paediatric population \cite{Bwakura2012}. Consequently, there has been a recent demand for incorporating more children into randomised controlled trials \cite{Caldwell2004} so drug efficiency and safety can be directly evaluated for children. In addition to being able to assess the efficiency of drugs within the paediatric population and determine suitable dosages, there are also advantages for the children enrolled in trials such as access to new medicine that may reduce morbidity. However, there are also many downsides including both physical and mental discomfort, separation from parents \cite{Caldwell2004} and the standard risks associated with adult clinical trials \cite{Lidt2004}. These downsides may be magnified in children due to potential errors in initially estimating `adult equivalent' doses, and due to additional impacts of drugs on still-developing tissues. In the worse case, a clinical trial could result in child mortalities. Therefore, if possible, it is more preferable to develop alternative means to identify ADRs that do not have these negative effects. 

Pharmacovigilance is the study of prescription drug ADRs, including the collection of suitable data and their detection and prevention. As the clinical trial data for the paediatric population is lacking, a key resource for paediatric pharmacovigilance is the spontaneous reporting system database \cite{Avery2011,Aagaard2013}. The spontaneous reporting systems amalgamate the suspected cases of ADRs that are voluntarily reported within a population. For example, in the UK if a patient is prescribed a drug and experiences an unexpected medical event then their doctor or the patient can report the suspected ADR via the yellow card scheme by filling out a form. All the yellow card scheme reports are then combined into a spontaneous reporting system database that is used to identify ADRs. As the database only contains reports detailing cases when a drug is taken and a suspected ADR occurs, and not the cases when a drug is taken and no ADRs is suspected, the frequency that drugs are prescribed is unknown. Furthermore, as the reporting is done voluntarily, it is common for data to be missing, incorrect or duplicated \cite{Bate2009}. Consequently, there are no current algorithms that can be applied to spontaneous reporting system databases that are capable of detecting ADRs with a high accuracy, nor are the algorithms able to quantify the frequency that the ADRs occur. It is the combination of a lack of clinical trials coupled with the general SRS limitations that makes the paediatric population potentially more susceptible to ADRs. Research into developing novel algorithms that are able to efficiently and effectively discover qualitative and quantitative ADR information for the paediatric population is required to reduce child morbidities and mortalities.

A new type of database, called the longitudinal observational database, has started to emerge as a potentially new source of ADR information. The longitudinal observational database contains sequences of patients' medical data often spanning decades of years and offers a new perspective for ADR discovery. One example of a longitudinal observational database is The Health Improvement Network (THIN) database that contains medical records extracted directly from general practitioners' databases across the UK (www.thin-uk.com). The THIN database contains validated personal information about each patient in the database including their year of birth and gender. There is also a complete medical record and prescription history for each patient during the time they have been registered at the practice. However, patients may not inform their doctors of all the medical events they experience, especially minor ones, and use of over-the-counter medication for self-treatment is unlikely to be recorded. As the THIN database contains information about how many patients are prescribed a drug, it may be possible to extract quantitative information about ADRs. The effect of dosage and frequency of prescription could also be investigated to identify the optimal treatment for each child. This knowledge could then be used to help personalise medication for paediatric patients based on their medical state, age and gender. 

Several methods have been presented to identify ADR signals using longitudinal observational databases, although comparisons have concluded that the methods generally have a high false positive rate \cite{Ryan2012,Reps2013a}. These algorithms include cohort techniques \cite{Jin2010}, case-series approaches \cite{Simpson2011}, case-control approaches \cite{harpaz2012}, disproportionality analysis \cite{Zorych2013} or a mixture of the previously mentioned techniques \cite{Noren2010}. In addition to being limited by a high false positive rate, many of these algorithms require the use of the patient's medical history years prior to the drug prescription of interest and this limits their ability for use on the paediatric population as a long medical history is often not available (if a child is very young). The case control method compares the prevalence of the drug within the population of patients experiencing the medical event and the prevalence of the drug within a population of patients that have similar covariates but do not experience the medical event. However, the case control technique applied to the paediatric population is likely to introduce confounding as children experiencing a medical event may have a serious disorder that makes them susceptible to other illnesses, whereas children not experiencing the medical event could be very healthy. Similarly, comparing children who take a drug with children not taking a drug may introduce confounding by indication \cite{Greenland1980} as children who do not take any medication are likely to be very healthy due to a lack of chronic or degenerative illnesses whereas children who require certain medication are likely to be very unhealthy. It follows that a novel method that does not suffer from confounding by indication or require long periods of historic medical history should be more effective at signalling ADRs.

Recently, a supervised framework \cite{Reps2013b} that learns from known ADRs has been proposed to signal ADRs by using features based on the Bradford Hill causality criteria, criteria that are frequently used to determine causality. This framework was shown to perform well, and obtained a low false positive rate even when the frequency of the ADR was low. Unfortunately, the calculation of many of the Bradford Hill features require knowledge of a long medical history and this is often not available for paediatric patients. Consequently, the framework using Bradford Hill criteria based features faces difficulties when applied to detect ADRs within the paediatric population. One possible solution would be to implement an analogous framework that uses features based on the counterfactual method for causal inference, as these features can be chosen such that a large medical history is not required. This may enable rapid detection of paediatric ADRs and the framework may yield a low false positive rate. The main benefit of this method is that it does not have the risks associated with clinical trials.

In this paper we aim to investigate whether an ensemble of specifically chosen simple study designs can signal acute ADRs within the paediatric population with a low false positive rate. A comparison will be implemented to determine whether the ensemble offers an improvement over each individual simple study design. The ensemble requires generating multiple distinct measures of association between a drug and medical event based on the counterfactual method for causal inference. However, each measure of assocaition will be chosen such that a distinct type of main confounding effect will be introduce. The motivation of combining these measures of association via an ensemble is that it may be possible to exploit any causal mechanism structures. Using drug and medical event pairs definitively known to represent ADRs or non-ADRs, the various measures of association are calculated for each drug and medical event pair to generate the labelled data used to train a random forest classifier. The measures of association for any drug and medical event pair with an unknown ADR status can then be calculated and fed into the trained random forest to determine whether the pair corresponds to an ADR. 

The objective of this paper is to investigate whether an ensemble of simple study designs can be implemented to signal acutely occurring side effects effectively within the paediatric population by using historical longitudinal data. The majority of pharmacovigilance techniques are unsupervised, but this research presents a supervised framework.

\section{Material} \label{sec:thin}
The THIN database contains temporal medical and therapy records for over 11 million UK patients. We used a subset of the THIN database for this research which contained records for 4 million patients.  Within the subset, there were a total of 30191726 medical events recorded for 1.7 million patients when they were 17 years old or less. For each patient, their year of birth, gender and other personal details are known. Each medical record specifies the patient that the record corresponds to, the record date and the medical event experienced by the patient. The medical events have a tree structure, with the medical event becoming more specific as its node depth (i.e., the length of the path from the root to the node) increases. Therefore, medical event nodes with a depth of 1 are the most general and medical event nodes with a depth of 5 are the most specific. 

Each prescription within the THIN database contains details about the specific drug prescribed and contains a code corresponding to the drug known as the British National Formulary (BNF) code \cite{bnf}. The BNF code has a hierarchal structure that can be used to identify similar drugs. For example, BNF codes starting with 05 (e.g., 05-xx-xx-xx) correspond to drugs used to treat infections, and BNF codes starting with 05-01-01 (e.g., 05-01-01-xx) correspond to penicillins. A drug family is the set of drugs with the same BNF code. For example, the drug family benzlypenicillin sodium and phenoxymethlypenicillin have a corresponding BNF code of 05-01-01-01 the drug family penicillinase-resistant penicillins have a BNF code of 05-01-01-02 and the drug family broad-spectrum penicillins have a BNF code of 05-01-01-03. These are the three drug families used to evaluate the novel framework presented in this paper. The number of records for each of the drug families in the THIN database is presented in Table \ref{drugs}. These drug families were chosen as they are frequently prescribed so their ADRs are generally well known and the creation of a reference set of definitive ADRs and non-ADRs was possible. 

\begin{table}\centering
\caption{Details about the records within the subset of the THIN database for a selection of three penicillin drugs prescribed to patients aged 17 years or less.}
\label{drugs}
\begin{tabular}{p{0.35\textwidth}p{0.2\textwidth}p{0.07\textwidth}p{0.18\textwidth}}
Drug Family & BNF & \multicolumn{2}{c}{Number of Prescriptions} \\
& & Total & First in 3 months \\ \hline \hline
benzlypenicillin sodium and phenoxymethlypenicillin & 05-01-01-01 &1520866 & 456926 \\
penicillinase-resistant penicillins & 05-01-01-02 & 310622 & 252947 \\
broad-spectrum penicillins& 05-01-01-03 & 6490455 & 1448563 \\
\end{tabular}
\end{table}

\section{Methodology} \label{sec:4}
\subsection{Ensemble of Simple Studies Design Framework Overview}
The proposed Ensemble of Simple Studies Design (ESSD) framework for signalling the acute ADRs that occur within the paediatric population is:

\begin{enumerate}
\item Generate simple studies labelled data

\begin{itemize}
\item Choose $n$ drug families of interest, where each drug family is denoted by $D_{k}, k \in[1,n]$.
\item Determine the risk medical events ($RME_{D_{k}}$), these are all the medical events that are potential acutely occurring ADRs to the drugs in $D_{k}$. 
\item For each drug family $D_{k}$ and medical event $\in RME_{D_{k}}$ pair, determine whether the medical event is a known ADR or non-ADR of the drug family $D_{k}$ and add labels. The label for the i\textsuperscript{th} pair is denoted by $y_{i}$. For example, if the i\textsuperscript{th} pair corresponds to an ADR then $y_{i}=1$ but if the i\textsuperscript{th} pair corresponds to a non-ADR then $y_{i}=0$.
\item Generate the features for each pair ($D_{k}$+event$\in RME_{D_{k}}$) by applying the simple study designs to calculate multiple estimated causal effect values (the measure of association). The feature vector for the i\textsuperscript{th} pair is denoted by $\mathbf{x_{i}}$.
\end{itemize}

\item Train a random forest model using the labelled data ($\{ (\mathbf{x_{i}}, y_{i}) \}$)
\begin{itemize}
\item Apply 20-fold cross validation to tune the random forest classifier.
\item Select the optimal model parameter by considering the classifiers general ability to rank pairs corresponding to ADRs above pairs corresponding to non-ADRs.
\end{itemize}

\item Apply the trained random forest classifier to the simple study design features of any unlabelled drug family and medical event pair (not known to correspond to an ADR or non-ADR) and classify the pair as an ADR or non-ADR.
\end{enumerate}

\subsubsection{Risk Medical Events}
For each drug family $D_{k}$, the medical events investigated are determined using temporal information. As we are interested in acutely occurring ADRs, we restrict our attention to only investigate medical events that are observed within the month after a prescription of any drug within $D_{k}$ is first prescribed. A month was chosen to be a suitable trade off to enable a sufficient amount of time for the patient to report the medical event while not introducing a surplus quantity of noise. Therefore, given a drug family $D_{k}$, the risk medical events of $D_{k}$ ($RME_{D_{k}}$) are defined as the set of all medical events that are observed for a minimum of three patients within the month after any prescription of a drug within $D_{k}$. We chose to add a limit of three or more patients experiencing the potential ADR as it is difficult to determine whether a medical event is an ADR if it is experienced by less than three patients. 

\subsubsection{Generating Features}
For each drug family ($D_{k}$) and medical event $\in RME_{D_{k}}$ pair, we extract six different estimates of the causal effect of $D_{k}$ on the medical event. These are the six simple study designs. The target population is the patients prescribed $D_{k}$ and the etiological time period (the period we investigate) is the month after the first prescription. The estimates of the causal effect are calculated by either using a different target population (Target Substitution) or using a different etiological time period in (Etiological Substitution). 
\begin{description}
\item[$x^{1}$: Etiological Substitution (SSD$_{1}$)-] The causal effect is approximated by comparing the risk of the medical event during the month after the prescription for the target population with the risk during the month before the prescription for the target population. The main confounding effect is caused by a covariate of `medical state' as the target population medical states are likely to change between the month before and the month after the prescription. This causal effect estimate is likely to be large for progressive medical events (e.g., progressions of the cause of taking the drug) even though they are not caused by the prescription. 

\item[$x^{2}$: Etiological Substitution (SSD$_{2}$)-] The causal effect is approximated by comparing the risk of the medical event during the month after the prescription for the target population with the risk during the year after for the target population. The main confounding effect is a covariate of `medical sate', but unlike $x^{1}$ the causal effect estimate is likely to be small for progressive medical events that become more common as the population ages and large for medical events that occurred acutely after the drug. 

\item[$x^{3}$: Target Substitution (SSD$_{3}$)-] The causal effect is approximated by comparing the risk of the medical event during the month after the prescription for the target population with the risk during a randomly chosen month for a substitute population that matches the target population on age and gender. The main confounding effect is caused by a covariate of `indication' as the target population all have certain illnesses causing them to require the drug but the substitute population does not. This causal effect estimate is likely to be large for medical events linked to the indication (i.e., the cause of taking the drug). 

\item[$x^{4}$: Target Substitution (SSD$_{4}$)-] The causal effect is approximated by comparing the risk of the medical event during the month after the prescription for the target population with the risk during the month after for a substitute population that are given a similar drug (i.e., a similar BNF code) and have the same indications. The main confounding effects are caused by a covariate of `medical caution' as patients may be prescribed different drugs for the same indication due to medical cautions (i.e., one patient has kidney issues preventing them from having the standard medication) or covariates of `age and gender' as age and gender may influence the choice of drug prescribed. 

\item[$x^{5}$: Etiological Substitution (SSD$_{5}$)-] First, a mapping is performed to `generalise' the medical event descriptions. This is done by mapping medical event nodes that have a depth greater than 3 to their depth 3 `parent node'. The causal effect is then approximated using the mapped data by comparing the risk of the corresponding depth 3 `parent node' medical event during the month after the prescription for the target population with the risk during the month before for the target population. This causal effect measure is less vulnerable to covariates of `medical event recording', where different medical events that correspond to the same/similar illness can be recorded due to redundancy or illness progression. 

\item[$x^{6}$: Etiological Substitution (SSD$_{6}$)-] First, a mapping is performed to `generalise' the medical event descriptions. This is done by mapping medical event nodes that have a depth greater than 4 to their depth 4 `parent node'. The causal effect is then approximated by comparing the risk of the corresponding depth 4 `parent node' medical event during the month after the prescription for the target population with the risk during the month before for the target population. This causal effect measure is less vulnerable to a covariate of `medical event recording', where different medical events that correspond to the same/similar illness can be recorded due to redundancy or illness progression. 
\end{description}

The vector ${\mathbf x}_{i}=(x_{i}^{1}, x_{i}^{2},...,x_{i}^{6})$ contains the six estimates of the causal effect for the $i$ \textsuperscript{th} drug family and medical event pair. We also create three additional features from the original,
\begin{equation*} 
x_{i}^{7} = \left\{ 
\begin{array}{l l}
x_{i}^{1}/x_{i}^{2} & \quad \text{if $|x_{i}^{2}|>0$ }\\
x_{i}^{1}& \quad \text{else}
\end{array} \right.
\end{equation*}
\begin{equation*} 
x_{i}^{8} = \left\{ 
\begin{array}{l l}
x_{i}^{1}/x_{i}^{4} & \quad \text{if $|x_{i}^{4}|>0$ }\\
x_{i}^{1}& \quad \text{else}
\end{array} \right.
\end{equation*}
\begin{equation*} 
x_{i}^{9} = \left\{ 
\begin{array}{l l}
x_{i}^{1}/x_{i}^{5} & \quad \text{if $|x_{i}^{5}|>0$ }\\
x_{i}^{1}& \quad \text{else}
\end{array} \right.
\end{equation*}
These additional features indicate how much the simple study design association measures deviate when considering time, similar patients or the specificity of the medical event. So the complete feature vector for each drug family and medical event pair is ${\mathbf x}_{i}=(x_{i}^{1}, x_{i}^{2},...,x_{i}^{9}) \in \mathbb{R}^{9}$.

\subsubsection{Random Forest Classifier}
A random forest is a supervised classifier. The task of supervised learning is to use the training data to learn a mapping between the feature vector and the class.  This mapping can then be used to predict the class for unseen data. The random forest is known as a ensemble classifier as it trains and combines weak and diverse classifiers. Each weak classifier is a decision tree that uses a subset of the available features (the simple design study measures of association) to predict the class (ADR or non-ADR). The advantages of the random forest classifier is that it can have features that are both discrete and continuous and does not require the features to be pre-processed (e.g., centred and scaled). The parameter of the random forest that needs to be chosen is the number of features that each decision tree can use, this is referred to as $mtry$. In this research we used the R implementation of the random forest \cite{randfor}. 

The random forest is trained using the labelled drug family and medical event pair data, $X^{L}= \{({\mathbf x}_{i}, y_{i}) |$ the label is known $ \}$. We applied 20-fold cross validation, this means that the data are partitioned into 20 sets and for each set the random forest is trained on the other nineteen sets and then apply to predict the class of each data-point within the set. The trained random forest is then evaluated by a user defined criteria, in our case the area under of receiver operating characteristic curve (AUC), and the average value corresponding to this measure over the twenty sets determines how well the random forest has performed. This performance measure is used to select the parameter $mtry$, as various random forests are train with difference values of $mtry$ and the $mtry$ that results in the highest AUC is selected. When we refer to the trained random forest we mean the random forest that has been trained using the $mtry$ value that was optimal.

\subsection{Evaluation}
To evaluate the ESSD framework, we created a reference set of drug families and medical events pairs that are known to be ADRs or non-ADRs. The drug families used to create the reference set are the penicillins with the BNF codes 05-01-01-01, 05-01-01-02 and 05-01-01-03. The reference is used to train the random forest and evaluate it. The reference set data corresponding to two of the drug family is used as labelled data to train the random forest and the reference set data corresponding to the remaining drug family is used for evaluation. 

\subsubsection{Creating the Reference Set}
The reference set was created by investigating all the $RME_{D_{k}}$ for each $D_{k}$. Medical events that are listed as side effects on the BNF website or the medical event states the occurrence of a adverse event or the medical event is candidiasis as antibiotics are known to cause this were labelled as ADRs. Any medical event with a cause that is known and is not related to the penicillins (e.g., impetigo, worms, diabetes) was labelled as non-ADRs and a selection of medical events that are likely to be related to the cause of taking the drug were also labelled as non-ADRs. The labels corresponding to medical events that are likely to be related are most likely to be incorrect out of all the labels as it is difficult to show that a medical event is not an ADR, as even events that cause the drug to be taken could also be ADRs. 

\subsubsection{Evaluation Measures}
The ESSD framework is evaluated by considering its ability to signal ADRs at its natural threshold and its ability to rank drug family and medical event pairs by how likely they correspond to ADRs. The ESSD is compared with each individual simple study design (SSD$_{i}$). 

The natural threshold evaluation measures are the number of true positives (TP), this is the number of pairs corresponding to ADRs that the method classes as ADRs and the number of false positives (FP) is the number of pairs corresponding to non-ADRs that the method classes as ADRs. Similarly, the number of false negatives (FN) is the number of pairs that correspond to ADRs but the method classes as non-ADRs and the number of true negatives (TN) is rhe number of pairs that correspond to non-ADRs and the method classes as non-ADRs. The natural threshold measures can be calculated as:
\begin{equation}\begin{split}
\mbox{Sensitivity} &= TP/(TP+FN) \\
\mbox{Specificity} &= TN/(FP+TN)\\
\mbox{False positive rate (FPR)} &= FP/(FP+TN)\\
\end{split}\end{equation}

The general ranking ability of the methods are measured by the average precision (AP) and the AUC. The AUC is the area under the curve of the sensitivity plotted against 1 minus the specificity for various thresholds. The AUC can be interpreted as the probability of a uniformly chosen ADR pair being ranked above a uniformly chosen non-ADR pair. If a method performs poorly at its natural threshold but has a high AUC, then this may mean the natural threshold needs to be modified.

\section{Results} \label{sec:5}
The ESSD framework was evaluated three times. Each time the reference set data for two of the drug familities were used to train the random forest and the reference set data for the third drug family was used for evaluation. A summary of each of the three evaluations is presented in Table \ref{tab:exp}. The optimal $mtry$ obtained when training the model is presented and also the result of the AUC for the cross-validation.

At their natural thresholds, the ESSD obtained a lower overall FPR of $0.149$ compared to the other methods that obtained FPRs between $0.184-0.716$. The ESSD had a sensitivity of 0.547.  Although other methods obtained a higher sensitivity, they also had a very high false positive rate ($\ge 0.532$) which is not desirable. The results of methods at their natural thresholds for each individual evaluation are presented in Table \ref{tab:nat_ab1} and the overall results with the sensitivity, specificity and false positive rate are displayed in Table \ref{tab:nat_ab2}.

The general ranking ability of the ESSD was consistently higher than the other methods with AUC values of 0.814, 0.806 and 0.813 for the evaluations 1 to 3 respectively. The highest AUC value out of all the other methods was 0.813, 0.794 and 0.729 for the evaluations 1 to 3 respectively. The AP value obtained by the ESSD was also greater for each evaluation, with the ESSD obtaining 0.615, 0.659 and 0.717 for evaluation 1 to 3 respectively, whereas the highest AP obtained by any other method over evaluation 1 to 3 was 0.587, 0.508 and 0.493 respectively. The results of the general ranking ability at presented in Table \ref{tab:gen_ab}.
\begin{table} \centering
\caption{The details of the evaluation experiments. }
\label{tab:exp}
\begin{tabular}{ccccc}
Evaluation & Training/testing Set & Optimal $mtry$ & Training AUC & Evaluation Set \\ \hline \hline
1 & $05-01-01-\{02,03\}$ & 4 & 0.885 & $05-01-01-01$ \\
2 & $05-01-01-\{01,03\}$ & 6 & 0.827 & $05-01-01-02$ \\
3 & $05-01-01-\{01,02\}$ & 9 & 0.875 & $05-01-01-03$ \\
\end{tabular}
\end{table}

\begin{table} \centering
\caption{For each evaluation experiment ,the number of TP, FP, FN and TN returned for the ESSD and each individual simple study design (SSD$_{i}$). }
\label{tab:nat_ab1}
\begin{tabular}{c cccc cccc cccc}
& \multicolumn{4}{c}{1} & \multicolumn{4}{c}{2} & \multicolumn{4}{c}{3}\\ 
method & TP & FP & FN & TN & TP & FP & FN & TN & TP & FP & FN & TN \\ \hline \hline
ESSD & 9 & 9 & 8 & 30 & 9 & 6 & 9 & 40 & 17 & 6 & 12 & 50\\
SSD$_{1}$& 17 & 21 & 0 & 18 & 15 & 31 & 3 & 15 & 26 & 33 & 3 & 23 \\
SSD$_{2}$& 17 & 22 & 0 & 17 & 18 & 42 & 0 & 4 & 29 & 37 & 0 & 19 \\
SSD$_{3}$& 5 & 4 & 12 & 35 & 5 & 13 & 13 & 33 & 17 & 17 & 12 & 39 \\
SSD$_{4}$& 2 & 4 & 15 & 35 & 2 & 17 & 16 & 29 & 4 & 5 & 25 & 51 \\
SSD$_{5}$& 17 & 19 & 0 & 20 & 17 & 26 & 1 & 20 & 25 & 30 & 4 & 26 \\
SSD$_{6}$& 17 & 19 & 0 & 20 & 15 & 29 & 3 & 17 & 24 & 31 & 5 & 25\\
\end{tabular}
\end{table}
\begin{table} \centering
\caption{The average number of TP, FP, FN and TN returned for the ESSD and each SSD$_{i}$ and the overall specificity, sensitivity and false positive rate (FPR). }
\label{tab:nat_ab2}
\begin{tabular}{c cccc ccc}
method & TP & FP & FN & TN & Sensitivity & Specificity & FPR \\ \hline \hline
ESSD & 35 & 21 & 29 & 120 & 0.547 & 0.851 & 0.149 \\
SSD$_{1}$& 58 & 85 & 6 & 56 & 0.906 & 0.397 & 0.603 \\
SSD$_{2}$& 64 & 101 & 0 & 40 & 1 & 0.284 & 0.716\\
SSD$_{3}$& 27 & 34 & 37 & 107 & 0.422 & 0.759 & 0.241\\
SSD$_{4}$& 8 & 26 & 56 & 115 & 0.125 & 0.816 & 0.184\\
SSD$_{5}$& 59 & 75 & 5 & 66 & 0.922 & 0.468 & 0.532 \\
SSD$_{6}$& 56 & 79 & 8 & 62 & 0.875 & 0.440 & 0.560 \\
\end{tabular}
\end{table}

\begin{table} \centering
\caption{A comparison of the general ADR ranking ability of the ESSD and each SSD$_{i}$. }
\label{tab:gen_ab}
\begin{tabular}{ccccccccc}
& \multicolumn{2}{c}{1} & \multicolumn{2}{c}{2} & \multicolumn{2}{c}{3} & \multicolumn{2}{c}{Average}\\ 
Method & AUC& AP & AUC & AP & AUC& AP & AUC & AP \\ \hline \hline
ESSD& 0.814& 0.615 & 0.806 & 0.659 & 0.813& 0.712 & 0.811 & 0.662 \\
SSD$_{1}$& 0.797& 0.559 & 0.606 & 0.362 & 0.678& 0.457 & 0.694 & 0.459 \\
SSD$_{2}$& 0.785& 0.548 & 0.741 & 0.430 & 0.729& 0.493 & 0.752 & 0.490 \\
SSD$_{3}$& 0.602& 0.499 & 0.484 & 0.292 & 0.629& 0.433 & 0.572 & 0.408 \\
SSD$_{4}$& 0.477& 0.316 & 0.322 & 0.217 & 0.459& 0.333 & 0.419 & 0.289 \\
SSD$_{5}$& 0.797& 0.558 & 0.794 & 0.508 & 0.698& 0.448 & 0.763 & 0.505 \\
SSD$_{6}$& 0.813& 0.587 & 0.621 & 0.356 & 0.655& 0.435 & 0.696 & 0.459 \\
\end{tabular}
\end{table}

\section{Discussion} \label{sec:6}
The results show that the ESSD framework consistently outperformed the individual simple study design measures. It consistently obtained a higher AP and AUC for all three BNF families investigated. At its natural threshold the ESSD framework was able to signal just over half the true ADRs while only signalling approximately 15\% of the non-ADRs. However, as it is difficult to prove a medical event is not an ADR, some of the non-ADR labels may be incorrect, so the probability of signalling an non-ADRs may actually be lower (i.e., the FPR is probably less than the value obtain within this research). The ESSD framework is also efficient, as the simple measures can be calculated readily and the classifier can be trained quickly. Once trained, the prediction is fast and could be implemented regularly when new data is added to the longitudinal database. As the ESSD has a low false positive rate and is efficient, it may be a useful framework to implement for signal generation. However, the signals that are generated will still need to be refined as it does not have a zero false positive rate.  

There have been few attempts to apply supervised learning to the field of pharmacovigilance but this is an interesting area, as the results of supervised classifier can be improved when more labelled data is available. Therefore, supervised methods should improve over time as as more ADR knowledge is gained. It may be possible to feedback the results of the supervised methodologies to further improve them via methods such as semi-supervised techniques. The ESSD also has the advantage of being adaptable, as it could incorporate new measures of association that get proposed over time as features.

In this paper we have trained the ESSD using similar BNFs to the BNF used for the evaluation. The justification for this is that similar BNF drugs are likely to have similar underlying patterns. It would be interesting to see whether the random forest within the ESSD framework could be trained on a variety of different BNFs and still perform well. This would indicate whether the underlying structures are independent of the drug. 

One difficulty that was noticed with the ESSD framework is the choice of BNF to use to calculate the similar BNF measure of association (SSD$_{4}$). For the penicillins that was easy as the BNF codes are numerous, but for BNF families such as the proton pump inhibitors, a closely resembling BNF family is difficult to determine and the wrong choice could lead to different models with various performances. To overcome this the methodology may need to be applied to individual drugs rather than BNF facilities, but this could be problematic when the drug is rarely prescribed and therefore rarely recorded within the database. However, with the combination of longitudinal healthcare databases now becoming common \cite{Overhage2012}, even rarely prescribed drugs should have a sufficient number of occurrences in the combined database.

\section{Conclusion} \label{sec:7}
In this paper we have proposed a novel framework, called the ensemble of simple study designs (ESSD), specifically for signalling acute ADRs within the paediatric population. The framework does not require knowledge of a patient's medical history of more than a month prior to the prescription. The results show that the ESSD can outperform each individual simple study design measures and appears to be more consistent. The ensemble still misclassifies some non-ADRs but it obtained a false positive rate of $0.149$, making it competitive with existing methods. The advantage of this methodology is that it is supervised, so as new ADRs are discovered its performance should increase as more labelled data will be available for training it.

Future work could involve researching new methods for refining the ADR signals that are generated and reducing the false positive rate further or investigating different features based on alternative substitutions for the hypothetical counterfactual situation. For example, features that deal with alternative forms of confounding such as the time of the year could be incorporated. It is also of interest to see whether adding more complex study design methods such as Temporal Pattern Discovery \cite{Noren2010} or HUNT \cite{Jin2010} can improve the ensemble and a comparison between these methods and the ESSD would be useful.

\begin{acknowledgements}
This study was funded by the Department of Computer Sciences, University of Nottingham, UK. J.M. Reps, J.M. Garibaldi, U. Aickelin, D. Soria, J.E. Gibson and R.B. Hubbard have no conflicts of interest that are directly relevant to the content of this study.
\end{acknowledgements}

\bibliographystyle{spmpsci} 
\bibliography{refs}


\end{document}